\documentclass[12pt]{article}
\usepackage[utf8]{inputenc}
\usepackage{hyperref}
\usepackage{amsmath}
\usepackage{graphicx}
\usepackage[round,authoryear]{natbib}
\usepackage{parskip}
\usepackage{authblk}
\usepackage{booktabs}    
\usepackage{multirow}    
\usepackage{longtable}   
\usepackage{geometry}    

\title{Considerations Influencing Offense-Defense Dynamics From Artificial Intelligence}
\author[1,2]{Giulio Corsi}
\author[1]{Kyle Kilian}
\author[1]{Richard Mallah}
\affil[1]{Center for AI Risk Management and Alignment}
\affil[2]{Leverhulme Centre for the Future of Intelligence, University of Cambridge}
\date{\today}
\begin{document}

\maketitle

\section{Introduction}

The rapid advancement of artificial intelligence (AI) technologies presents profound challenges to societal safety. As AI systems become more capable, accessible, and integrated into critical services, the dual nature of their potential is increasingly clear. While AI can enhance defensive capabilities in areas like threat detection, risk assessment, and automated security operations \citep{RN2108}, it also presents avenues for malicious exploitation and large-scale societal harm, for example through automated influence operations and cyber attacks \citep{RN2110,RN2090}.

Understanding the dynamics that shape AI's capacity to both cause harm and enhance protective measures is essential for informed decision-making regarding the deployment, use, and integration of advanced AI systems. This paper builds on recent work on offense-defense dynamics within the realm of AI \citep{RN2119,RN2118}, proposing a taxonomy to map and examine the key factors that influence whether AI systems predominantly pose threats or offer protective benefits to society. By establishing a shared terminology and conceptual foundation for analyzing these interactions, this work seeks to facilitate further research and discourse in this critical area.

\textbf{Research Objectives:}
\begin{enumerate}
    \item Identifying and defining a taxonomy of key factors influencing the proliferation of offensive and defensive uses of AI relative to society.
    \item Exploring the implications of these dynamics for AI governance and policy.
\end{enumerate}

\section{Offense-Defense Dynamics in the Context of AI Impacts}

The concept of offense-defense dynamics originates from international relations and military strategy, where it is used to assess the likelihood of conflict and the strategic balance between opposing forces \citep{RN2111,RN2123}. In that context, the balance between offensive and defensive capabilities influences nations' decisions regarding aggression, deterrence, and arms development.

Applying this concept to AI involves examining the interplay between an AI system's capacity to cause societal harm and its ability to mitigate risks through defensive applications. There is a growing body of research exploring both offensive and defensive applications of AI in various domains, such as cybersecurity \citep{RN2108,RN2114}, influence operations \citep{RN2110,RN2115}, and CBRN (Chemical, Biological, Radiological, and Nuclear) threats \citep{RN2117,RN2116}. However, the overall balance and dynamics between societally offensive and defensive AI capabilities remain unclear and often contentious. This section introduces key concepts and definitions that underpin the application of offense-defense theory to AI.

\subsection{Prioritizing Societal Impacts}

In applying offense-defense theory principles to AI, we adopt a global framing that reflects the necessity of considering the widespread impacts of AI systems on society as a whole. Unlike traditional dyadic analyses that focus on the interactions between two specific entities, our approach positions society itself as the primary entity affected by AI. This global perspective is essential as the effects of releasing an advanced AI system are inherently global, influencing not just specific adversaries or defenders but the entire societal landscape. By treating society as the first party, we can factor it out of the dyadic equation and focus on how the effects of an AI model causally radiate from its release. This allows for the analysis of the offense-defense dynamics of a model in terms of its impact on societal structures, vulnerabilities, and defenses. 

In this context, we are not merely considering the offense-defense balances of individual models in isolation. Instead, we are examining how each model's offensive and defensive capabilities interact with societal factors to influence the overall balance of threats and protections within the global community. An AI system's potential for societal harm or benefit is determined by a complex interplay of factors, including its technical characteristics, the contexts in which it is deployed, and the ways in which its capabilities propagate through society.

Therefore, when discussing offense-defense dynamics in AI, it is crucial to prioritize the factors that enable the proliferation of offensive or defensive applications from a societal perspective. This approach acknowledges that the key metrics of an AI system's impact lie not only in its individual characteristics but in how its release shapes the broader landscape of technological capabilities and their societal consequences.

\subsection{Offense-Defense Neutrality}

In applying this framing to AI, it is important to recognize that advanced AI systems are inherently dual-use and generally agnostic to offensive or defensive orientations; the same underlying technologies can be harnessed---albeit with some notable exceptions---for both beneficial and harmful purposes \citep{RN2122,RN2101,RN2121,RN2131}. For instance, machine learning algorithms designed for pattern recognition can be employed to enhance cybersecurity by detecting anomalies, or conversely, to develop sophisticated cyber-attacks that evade detection \citep{RN2098}. This duality underscores the fact that the potential for offense or defense is not an intrinsic characteristic of the AI technology itself. Instead, it is shaped by how the technology is developed, deployed, and controlled within specific technical and contextual frameworks.

The orientation of an AI system towards societal offense or defense emerges from a complex interplay of sociotechnical factors that include attributes such as a system’s capabilities and adaptability, and contextual factors such as the regulatory environment and ongoing international tensions. A given AI system can also be offense-dominant within one application area or domain while simultaneously being defense-dominant in another application area or domain, and there are a myriad of domains and application areas. For these reasons, the present work does not concentrate on cataloging specific applications, which may be fluid and context-dependent, but rather, on using such applications to derive features and conditions that may facilitate the proliferation of societally offensive or defensive applications of AI. By focusing on these enabling factors, we aim to better understand and influence the overarching dynamics that determine offensive and defensive dynamics for AI systems.

\subsection{Asymmetries in AI Offense-Defense Dynamics}

While AI technologies may be fundamentally neutral, significant asymmetries can emerge naturally in their offensive and defensive applications. Similar to the lessons learned from cybersecurity, these asymmetries may emerge from differences in resource requirements, technological barriers, and strategic advantages inherent to offensive and defensive applications \citep{RN2124,RN2126,RN2125}. When applying offense-defense theory to AI, it becomes evident that offensive AI applications can often possess intrinsic advantages, being comparatively easier to develop and deploy.

The offensive advantage in AI applications stems from the fact that exploiting existing vulnerabilities typically requires less coordination and planning than comprehensive defense. For instance, the generation of disinformation through deepfake technologies can be accomplished with relatively accessible tools \citep{RN2128,RN2095,RN2106}, producing a spectrum of impacts and permeation levels that may be challenging to anticipate and counteract. This ease of offensive deployment contrasts sharply with the demands of defensive measures \citep{RN2099}. This balance will tend to be magnified exponentially as a risk surface grows exponentially.

Defensive applications, by their nature, often necessitate more sophisticated solutions, greater collaboration among stakeholders, and continual adaptation to evolving threats. Developing effective defenses against AI-driven offenses involves complex challenges such as detecting subtle anomalies, countering adaptive adversaries, and integrating protective measures across multiple systems and platforms. These requirements create higher barriers to entry and sustained efficacy for defensive AI applications. The naturally occurring asymmetries between offensive and defensive AI applications are likely to be a key element in the understanding and modeling of offense-defense dynamics for AI.

\subsection{Core Definitions}

Finally, to effectively analyze and discuss the offense-defense dynamics in AI, it is important to establish a common vocabulary. The following definitions provide a foundation for our taxonomy, allowing for consistent communication and analysis of the complex interplay between offensive and defensive applications of AI technologies. These terms will be used throughout the remainder of this paper to explore the factors influencing the societal impact of AI systems.

\begin{description}
    \item[Offensive AI Applications:] AI systems designed or utilized to conduct attacks, exploit vulnerabilities, or disrupt systems and societies. Such applications make society more at risk. Examples include automated hacking tools, deepfake technologies used in influence operations, autonomous weapon systems, and AI-augmented malware.
    
    \item[Defensive AI Applications:] AI systems designed or utilized to protect, detect, and respond to threats. These applications make societies safer. Examples include AI-driven fraud detection systems, AI-enhanced threat monitoring and anomaly detection, and Intrusion Detection Systems (IDS).
     
    \item[Offense-Defense Dynamics:] The complex interplay between offensive and defensive AI applications within a given sociotechnical context. This concept encompasses the relative ease of developing and deploying offensive versus defensive AI applications, the comparative effectiveness of offensive and defensive AI capabilities, and the societal impact of both types of applications.
\end{description}

\section{Offense-Defense Dynamics Taxonomy}

To systematically analyze the factors influencing the offensive and defensive applications of AI systems, we introduce the Offense-Defense Dynamics Framework. This taxonomy comprises six interrelated elements that collectively shape the proliferation and impact of offensive and defensive AI applications within a given context. These elements are derived from the bottom-up, analyzing known examples of defensive and offensive AI uses, and abstracting from these to identify general patterns and principles. By analyzing these elements, we aim to provide a structured approach to understanding how AI technologies can shift the balance between societal risk and safety.

The six elements that compose the taxonomy are:
\begin{enumerate}
    \item Raw Capability Potential
    \item Accessibility and Control
    \item Adaptability
    \item Proliferation, Diffusion, and Release Methods
    \item Safeguards and Mitigations
    \item Sociotechnical Context
\end{enumerate}

\subsection{Raw Capability Potential}

\textbf{Definition:} Raw Capability Potential refers to the inherent abilities of an AI system to perform actions that can be utilized either offensively or defensively, independent of any external safeguards or mitigations. It represents the fundamental power of the AI system to harm or protect societal assets, infrastructure, individuals, or the biosphere, based solely on its technical capacities. This concept focuses on what an AI system can do at a fundamental level, without considering its current usage or any protective measures in place.

\textbf{Importance in Offense-Defense Dynamics:} The raw capabilities of an AI system set the baseline for both its potential benefits and risks. Advanced capabilities enable sophisticated defensive measures, such as enhanced threat detection and response, but they also increase the potential for offensive applications, like automated cyber-attacks or the creation of disinformation at scale.

When assessing Raw Capability Potential, we currently consider two primary, high-level dimensions:

\begin{itemize}
    \item \textbf{Capabilities Breadth:} This refers to the range of tasks and domains an AI system can operate in, reflecting its versatility and expertise across domains and modalities. It encompasses the system's ability to function across multiple areas, from natural language processing to mathematical modeling. For example, a large language model capable of generating text, analyzing code, solving mathematical problems, and providing insights across various scientific disciplines demonstrates high capabilities breadth. Such a model could be used defensively to detect vulnerabilities across multiple domains or offensively to generate sophisticated multi-vector attack strategies \citep{RN2109,RN2121}.
    
    \item \textbf{Capabilities Depth:} This refers to the level of sophistication and effectiveness an AI system demonstrates within specific domains. It describes how advanced and capable a system is in its particular area of focus, regardless of how broad or limited that area might be. For example, an AI system specifically designed for protein folding prediction, such as AlphaFold \citep{RN2092}, exhibits high capability depth in its domain. While its application range might be narrow, this depth allows for a number of offensive and defensive uses, such as the discovery of new treatments \citep{RN2107}, and the design of harmful biological agents \citep{RN2093,RN2094}.
\end{itemize}

While both breadth and depth of capabilities can significantly impact offense-defense dynamics, it is important to recognize that these dimensions are not mutually exclusive. Rather, AI systems can exhibit various combinations of breadth and depth, producing multiple gradients of potential capabilities. Models can be simultaneously broad and deep, amplifying their potential influence on offense-defense balances.

\subsection{Accessibility and Control}

\textbf{Definition:} Accessibility and Control refer to who can access, use, and operate an AI system in its existing form, and the mechanisms that govern and restrict this access. It focuses on user authentication, licensing for use, and control mechanisms that determine who can interact with the AI system as it currently exists, without modification. This concept also considers whether other AI systems can access or interface with the system in question, which is increasingly relevant in scenarios involving multi-agent systems.

\textbf{Importance in Offense-Defense Dynamics:} The level of accessibility and control over an AI system significantly influences its potential for both beneficial use and misuse. Systems that are widely accessible, with minimal control and low barriers to entry, may be more susceptible to exploitation for offensive purposes. Conversely, restricted access can limit the spread of defensive applications but also reduces the risk of unauthorized use. Balancing accessibility and control is essential to maximize societal benefits while minimizing risks.

When assessing Accessibility and Control, we currently consider two high-level dimensions likely to influence offense-defense dynamics:

\begin{itemize}
    \item \textbf{Access Level:} This includes considerations on an AI system’s access level (public, restricted or confidential), access type (open source, proprietary with public API or closed system), and access cost (low, medium or high). For example, an open-source language model like LLAMA is highly accessible, being free and open-source. This high accessibility can facilitate its use in defensive applications, such as medical assistance and content moderation \citep{RN2096,RN2097}. However, it also lowers the barrier to offensive applications, such as generating malicious disinformation and exploiting privacy vulnerabilities \citep{RN2098,RN2101}. Additionally, disparities in access can influence the effectiveness of defensive measures while amplifying vulnerabilities. Entities or populations with limited access to AI—due to factors like low general knowledge, technical expertise, language barriers, or restricted internet connectivity—may be unable to leverage AI for defensive purposes. For instance, communities relying on limited internet services, such as those provided by balloon-based networks, might only receive AI-generated outputs without fully understanding the underlying technology, thereby reducing their capacity to counteract malicious AI-driven activities on platforms like social media.
    
    \item \textbf{Interaction Complexity:} This refers to the range and sophistication of ways users and systems can engage with the AI model within its existing configuration. It encompasses the spectrum of possible interactions, from simple query-response to complex multi-turn dialogues and task completion. It also includes the granularity of control in formulating inputs and interpreting outputs. For example, a model that only allows single-turn, text-based queries has lower interaction complexity compared to one that enables multi-modal inputs, multi-turn conversations, and fine-grained control over response parameters. Lower interaction complexity can limit potential misuse by restricting the ways users can leverage the model's capabilities, while higher complexity might enable more sophisticated applications but also increases the potential for unintended uses \citep{RN2099}.
\end{itemize}

\subsection{Adaptability}

\textbf{Definition:} Adaptability refers to the capacity of an AI system to be modified, customized, or repurposed beyond its original context or intended use. Key factors influencing adaptability include the accessibility of model weights and fine-tuning capabilities, the feasibility of model distillation or knowledge transfer, and the presence of modular components that can be independently updated or replaced.

\textbf{Importance in Offense-Defense Dynamics:} An AI system's adaptability significantly impacts its potential for both offensive and defensive applications. Highly adaptable systems can be swiftly repurposed for new tasks or domains, which can be advantageous for defensive measures by allowing rapid responses to emerging threats. However, this flexibility also presents risks, as malicious actors may exploit adaptable systems for harmful purposes. Moreover, the adaptability of AI systems influences the likelihood of unintended consequences. As systems are modified or repurposed, there's an increased potential for introducing new vulnerabilities or expanding the attack surface, which could shift the balance in offense-defense dynamics.

When assessing Adaptability, we currently consider two high-level dimensions likely to influence offense-defense dynamics:

\begin{itemize}
    \item \textbf{Modifiability:} This dimension encompasses the ease with which an AI system can be altered, including the feasibility of fine-tuning and the presence of modular and extensible architectures. These characteristics enable a model to be modified and repurposed beyond its initial scope. For instance, an open-source model with accessible weights and a modular structure exhibits a high degree of modifiability. This allows for fine-tuning the model for specific offensive and defensive applications, ranging from swift detection of malicious activities \citep{RN2100} to potentially generating harmful content or unethical advice \citep{RN2102,RN2105}.
    
    \item \textbf{Knowledge Transferability:} This aspect relates to the AI system's ability to transfer its learned knowledge and capabilities to new tasks, domains, or model architectures. It encompasses mechanisms such as model distillation and transfer learning. High knowledge transferability allows an AI system to be quickly adapted or compressed for new applications without extensive retraining. For instance, a model with strong distillation capabilities could be compressed into a lightweight version for edge device deployment, enhancing its adaptability for various defensive applications like real-time threat detection \citep{RN2137}. Conversely, transfer learning capabilities could enable rapid adaptation of a model to generate domain-specific content \citep{RN2135,RN2136}. The ease of knowledge transfer significantly influences an AI system's overall adaptability, affecting its potential impact on offense-defense dynamics across different scenarios.
\end{itemize}

\subsection{Proliferation, Diffusion, and Release Methods}

\textbf{Definition:} This element refers to the strategies and mechanisms through which an AI model can be distributed, shared, and disseminated across society. It focuses on the strategies and pathways through which AI systems become available and are integrated into various contexts.

\textbf{Importance in Offense-Defense Dynamics:} The manner in which AI systems are released and proliferate affects how quickly and widely they can be adopted for both offensive and defensive purposes. Open and uncontrolled release methods can lead to rapid diffusion, increasing the potential for misuse, while controlled release can limit access to authorized users but may also slow down beneficial adoption. Assessing proliferation and diffusion methods is key for understanding the positive and negative proliferation dynamics of AI technologies in society.

When assessing Proliferation, Diffusion, and Release Methods, we currently consider two high-level dimensions likely to influence offense-defense dynamics:

\begin{itemize}
    \item \textbf{Distribution Control:} This dimension includes the release method (such as fully open-source release, open weights release, or closed-source release) and the degree of control exerted over distribution. For instance, the phased release strategy employed for GPT-4, where access was initially restricted to approved developers and gradually expanded, exemplifies high distribution control. This approach enables careful monitoring and adjustment of the model's impact, potentially mitigating early risks and vulnerabilities. In contrast, the release of BLOOM, a large language model with openly available weights and code, demonstrates low distribution control. While this open approach may facilitate research into AI models, it may also heighten the risks of misuse or unintended applications \citep{RN2095}.
    
    \item \textbf{Model Reach and Integration:} This dimension involves the ease with which an AI model can be deployed across various platforms and integrated into existing systems. It encompasses factors such as the model's size, compatibility with different hardware, and interoperability with various software ecosystems. Models with high reach and integration can be rapidly deployed across diverse geographical regions and easily incorporated into a wide range of applications. For instance, a lightweight language model that can run efficiently on mobile devices and integrate seamlessly with popular messaging platforms would have high reach and integration. This characteristic can accelerate the adoption of AI for beneficial purposes, such as real-time language translation or content moderation. However, it also increases the potential for widespread misuse, enabling malicious actors to quickly deploy AI for tasks like generating personalized phishing messages or creating large-scale disinformation campaigns. For example, GPT-J, an open-source language model, demonstrates a high deployment reach and integration due to its complete availability, relatively compact size (6 billion parameters), and ease of deployment. This rapid dissemination has enabled various potential misuses, including the creation of WormGPT, a language model designed for cybercriminal activities \citep{RN2104}.
\end{itemize}

\subsection{Safeguards and Mitigations}

\textbf{Definition:} Safeguards and Mitigations encompass a comprehensive set of measures implemented to prevent misuse and limit potential harm from AI systems. This includes technical safeguards embedded within the AI models themselves, ethical guidelines governing their development and use, stringent usage policies, and robust monitoring and auditing systems. These measures are designed to ensure responsible use of AI technologies while maximizing their beneficial impact.

\textbf{Importance in Offense-Defense Dynamics:} Effective safeguards and mitigations can significantly reduce the risk of AI systems being exploited for offensive purposes while enhancing their defensive utility. Implementing strong safeguards ensures that AI technologies contribute positively to societal safety and align with ethical standards. Understanding these measures is crucial for policymakers and developers aiming to mitigate risks associated with advanced AI systems.

When assessing Safeguards and Mitigations, we consider two main taxonomy elements:

\begin{itemize}
    \item \textbf{Technical Safeguards:} This includes content filtering mechanisms, behavioral alignment techniques, and output limitations. For example, models with robust technical safeguards that enable the refusal of harmful outputs, including the ability to decline requests for explicit content or potentially dangerous information, can significantly reduce the risk of misuse \citep{RN2105}. 
    
    \item \textbf{Monitoring and Auditing:} This aspect focuses on the ongoing processes and taxonomies established to oversee AI system usage, detect anomalies, and ensure compliance with ethical guidelines and legal requirements. It includes real-time usage monitoring practices, periodic safety audits, and the implementation of accountability taxonomies. For instance, AI models that undergo rigorous safety audits prior to public release, coupled with controlled API access featuring robust output monitoring, can significantly mitigate emerging threats. This approach allows for early detection and prevention of potential misuse, enhancing the overall security of the AI system \citep{RN2132,RN2133}.
\end{itemize}

\subsection{Sociotechnical Context}

\textbf{Definition:} The Sociotechnical Context refers to the broader technological, social, and political environment in which the AI system operates and interacts. It includes factors such as regulatory frameworks, public awareness, geopolitical tensions, and the robustness of technological infrastructure.

\textbf{Importance in Offense-Defense Dynamics:} The sociotechnical context can either amplify or mitigate the risks associated with AI technologies. A supportive environment with robust regulations, international cooperation, and high public awareness can enhance defensive applications and promote responsible use of AI. Conversely, a hostile context marked by geopolitical tensions, ineffective regulations, and low public awareness can increase the likelihood of proliferation of offensive applications and misuse of AI technologies. Understanding the sociotechnical context is essential for stakeholders to develop appropriate strategies and policies.

When assessing Sociotechnical Context, we consider two main dimensions:

\begin{itemize}
    \item \textbf{Geopolitical Stability:} This includes factors such as international cooperation, conflicts and tensions, and global power dynamics in AI. The geopolitical environment influences motivations for developing offensive or defensive AI capabilities and affects collaboration on AI governance. For example, increasing tensions between major powers in AI development, such as the US and China, could lead to an arms race scenario, potentially prioritizing offensive applications over safety considerations \citep{RN2106,RN2134}.
    
    \item \textbf{Regulatory Strength:} This pertains to the maturity and effectiveness of regulatory taxonomies on AI and the capacity for enforcement. Strong regulations and enforcement can mitigate risks by setting standards for responsible AI development and use, while weak regulations may allow harmful applications to proliferate. For example, the European Union's proposed AI Act aims to establish a comprehensive regulatory taxonomy for AI systems, potentially enhancing safety and ethical standards while possibly slowing down dangerous and offense-dominant AI applications compared to regions with less stringent regulations \citep{RN2130}.
\end{itemize}

\subsection{Framework Interactions}

Notably, the six elements of the Offense-Defense Dynamics Framework form a complex, interconnected system rather than existing in isolation. This interconnectedness is crucial for a comprehensive understanding of AI's offense-defense dynamics and for developing effective policies and strategies. Each element---Raw Capability Potential, Accessibility and Control, Adaptability, Proliferation, Diffusion, and Release Methods, Safeguards and Mitigations, and Sociotechnical Context---influences and is influenced by the others in a web of dynamic relationships.

These interdependencies manifest in various ways throughout the taxonomy. For example, an AI system's level of access control significantly impacts proliferation patterns, as open-source models with minimal restrictions tend to diffuse more rapidly than closed, proprietary systems. Similarly, the regulatory environment and geopolitical climate, encapsulated in the Sociotechnical Context, exert a pervasive influence across all other elements. They shape the speed and manner of AI proliferation, the development of capabilities, and the implementation of safeguards. Further, highly adaptable systems, while offering flexibility for various applications, may pose challenges for implementing effective safeguards, as they can be more easily modified to circumvent protective measures.

Additionally, within this interconnected model, certain elements emerge as clear leverage points, exerting disproportionate influence on the overall offense-defense dynamics. Accessibility and Control, for instance, can serve as a critical bottleneck; regardless of an AI system's raw capabilities or adaptability, tightly controlled access can significantly limit its potential for both offensive and defensive applications. Similarly, robust Safeguards and Mitigations can act as a bottleneck for offensive applications, potentially neutralizing threats even from highly capable and adaptable systems. The Sociotechnical Context, through its influence on policy decisions and international cooperation, can become a significant leverage affecting all other elements.

Understanding these complex interactions is essential for the analysis of AI's potential societal impacts. It highlights the need for a holistic, systems-level approach to managing offense-defense dynamics in AI, where changes in one element are considered in light of their potential ripple effects throughout the entire taxonomy. This perspective can inform more effective strategies for maximizing the defensive potential of AI while mitigating its offensive risks, ultimately contributing to a safer and more beneficial integration of AI into society.

\newpage
\subsection{Applying the Offense-Defense Dynamics Taxonomy: The Use of AI to Generate and Detect Disinformation} \label{sec:taxonomy_implications}

To demonstrate the practical application of the Offense-Defense Dynamics Taxonomy, we present an illustrative example examining how AI models capable of generating multimodal content can be used both to produce and to detect disinformation and coordinated influence operations. The following explores the implications of each taxonomy element for offensive and defensive applications of AI within information environments.
\vspace{5pt}

\begin{longtable}{@{}p{3cm}p{3cm}p{9cm}@{}}
\toprule
\textbf{Taxonomy Component} & \textbf{Subcomponent} & \textbf{Implications for AI and Disinformation} \\ \midrule
\endhead

\multirow{2}{3cm}{1. Raw Capability Potential} 
    & \textbf{Capabilities Breadth} 
    & AI models with broad capabilities can generate disinformation across various domains and areas of expertise. This versatility allows for the creation of sophisticated disinformation campaigns that can target different audiences. Conversely, the same breadth can enable the development of robust detection tools that may enhance the detection and countering of disinformation. \\ 
    \cmidrule(l){2-3}
    & \textbf{Capabilities Depth} 
    & Models with deep expertise in specific domains can create highly credible and tailored disinformation. For example, an AI system with deep knowledge in medical science can generate misleading health information that appears authentic to both laypersons and professionals. This depth increases the potential harm, as the disinformation can be more persuasive and harder to debunk, potentially influencing public opinion or behaviors in critical areas like health, finance, or politics. On the other hand, this depth of expertise can be harnessed to enhance detection mechanisms, allowing for the identification of subtle inaccuracies in specialized content, supporting efforts to counter disinformation within specific domains. \\ \midrule

\multirow{2}{3cm}{2. Accessibility and Control} 
    & \textbf{Access Level} 
    & High accessibility to powerful AI models lowers the barrier for malicious actors to generate disinformation. Open-source models or those with minimal cost and restrictions enable widespread misuse. For example, if a state-of-the-art language model is publicly available without stringent controls, anyone can leverage it to produce and disseminate disinformation at scale. \\ 
    \cmidrule(l){2-3}
    & \textbf{Interaction Complexity} 
    & Advanced interaction capabilities, such as multi-turn conversations and fine-grained control over outputs, allow users to craft more sophisticated and targeted disinformation. For instance, an AI system that accepts detailed prompts can be guided to generate specific narratives or mimic particular writing styles, making the disinformation more convincing. Lower interaction complexity might limit misuse by restricting the level of customization, but it could also reduce the utility of the AI for legitimate applications. \\ \midrule

\multirow{2}{3cm}{3. Adaptability} 
    & \textbf{Modifiability} 
    & Highly modifiable AI systems can be fine-tuned or altered to bypass content filters and safeguards, enabling the generation of disinformation that the original model might restrict. For example, malicious actors could fine-tune an open-source model on datasets containing biased or false information to enhance its ability to produce persuasive disinformation. Conversely, this adaptability may allow defenders to customize models to better detect and counter emerging disinformation tactics by promptly updating detection systems. \\ 
    \cmidrule(l){2-3}
    & \textbf{Knowledge Transferability} 
    & AI models that allow for easy knowledge transfer can have their capabilities distilled into smaller models or transferred to different platforms, increasing the reach of disinformation tools. For instance, a powerful language model could be distilled into a lightweight version deployable on mobile devices, facilitating decentralized disinformation campaigns that are harder to monitor and control. This transferability exacerbates the spread and impact of AI-generated disinformation across various channels and devices. \\ \midrule

\multirow{2}{3cm}{4. Proliferation, Diffusion, and Release Methods} 
    & \textbf{Distribution Control} 
    & The method by which AI models are released significantly influences their potential misuse in disinformation campaigns. Unrestricted, open releases---such as publishing model weights without any access controls---enable anyone to utilize these tools for generating disinformation, making it challenging to track or prevent malicious activities. For example, a publicly released model might be downloaded and used to produce propaganda at scale. Controlled distribution methods, like providing access through monitored APIs with usage policies and oversight, can mitigate this risk by allowing developers to detect and respond to misuse. \\ 
    \cmidrule(l){2-3}
    & \textbf{Model Reach and Integration} 
    & The ease with which AI models can be integrated into existing platforms and applications magnifies their impact on disinformation dissemination. Models designed with interoperability in mind can be seamlessly embedded into social media platforms, messaging apps, or content management systems. For instance, AI-powered bots using such models could generate and distribute false narratives across multiple channels, reaching large audiences rapidly and potentially influencing public discourse. High integration potential not only accelerates the spread of disinformation but also makes detection and intervention more complex, as disinformation becomes interwoven with legitimate content. \\ \midrule

\multirow{2}{3cm}{5. Safeguards and Mitigations} 
    & \textbf{Technical Safeguards} 
    & Implementing robust technical safeguards within AI models is crucial for preventing their misuse in generating disinformation. Techniques such as content filtering, ethical alignment during training, and response moderation can significantly reduce the AI's ability to produce harmful content. For example, incorporating reinforcement learning from human feedback (RLHF) can guide the model to avoid generating false or misleading information. \\ 
    \cmidrule(l){2-3}
    & \textbf{Monitoring and Auditing} 
    & Continuous monitoring and auditing of AI systems are vital for early detection and prevention of disinformation generation. By analyzing usage patterns, developers and stakeholders can identify anomalies indicative of misuse, such as unusually high volumes of content generation or requests involving sensitive topics. For instance, monitoring API calls that frequently attempt to produce content violating terms of service can help in taking proactive measures. Regular audits of the AI models and their outputs can ensure adherence to ethical guidelines and compliance with regulations. \\ \midrule

\multirow{2}{3cm}{6. Sociotechnical Context} 
    & \textbf{Geopolitical Stability} 
    & The geopolitical environment significantly influences the risks associated with AI-generated disinformation. In regions experiencing high tensions or conflicts, state and non-state actors are more likely to exploit AI technologies to influence public opinion, destabilize governments, or sow discord among populations. For example, during an election cycle in a politically unstable country, foreign entities might deploy AI-generated deepfake videos or fabricated news articles to sway voters or undermine confidence in the electoral process. The global nature of AI technology exacerbates this issue, as actors can operate across borders with relative anonymity. \\ 
    \cmidrule(l){2-3}
    & \textbf{Regulatory Strength} 
    & The presence of strong, enforceable regulations plays a crucial role in deterring the misuse of AI for disinformation. For instance, regulations requiring platforms to label AI-generated content or verify the authenticity of information sources can help users make informed judgments about the credibility of the content they consume. Effective enforcement mechanisms are essential; without them, regulations may have little practical impact. In regions with weak regulatory systems, malicious actors may exploit these gaps, operating with minimal risk of repercussions. Therefore, strengthening regulatory frameworks and ensuring their consistent application is vital for mitigating the spread and impact of AI-generated disinformation. \\ 

\bottomrule
\end{longtable}

\section{Implications for AI Governance and Policy}

AI offense-defense dynamics have several important policy implications, particularly in shaping regulatory frameworks around the development, deployment, and control of advanced AI systems. Developing a granular understanding of offense-defense dynamics can give policymakers a more detailed view of how and to what degree advanced AI systems might be leveraged for attacks, where critical vulnerabilities are most prominent, and areas to expend resources to ensure critical infrastructure or the information ecosystems remain secure (e.g., from malicious attacks or AI system failures). At a regulatory level, this core of a framework can highlight the risks of specific company policies, such as open model weights or open access, and how, combined with highly capable models, could lead to bad actors harnessing the most capable models or losing control.

The presented framing focuses on the interactions and interdependencies across AI system capabilities, access and diffusion, and control measures to highlight under which conditions advanced AI systems could favor offense over defense: A better understanding of these interactions could shift AI safety priorities, policy reporting requirements---such as those enacted in Executive Order (E.O.) 14110---and the development of new standards by government agencies, such as the Bureau of Industry and Security (BIS) and the National Institute of Standards and Technology (NIST). These outputs could help inform the network of AI Safety Institutes (e.g., U.S. or UK AISI) to fast-track interventions to protect the public.

A more detailed understanding of AI capabilities, their interactions, and their propensity to prevent, exacerbate, or cause societal harm can help international partners manage AI's offensive potential. Offensive AI applications, such as autonomous cyberattacks or disinformation campaigns, can be deployed relatively easily, with increasing scale and scope, and often with fewer resources than what is required for effective defense. This asymmetry presents a significant threat to global security, as adversarial actors may scale offensive efforts to exploit vulnerabilities in critical infrastructures with minimal effort. Policymakers must prioritize policy efforts for transparency, accountability, and control, particularly in areas with high offense dominance that could rapidly undermine state security, public trust, and economic stability. International AI agreements, similar to arms control treaties, may be necessary to prevent the unchecked proliferation of AI weapons and offensive capabilities.

On the defensive side, policymakers must also consider the higher resource and coordination requirements for effective AI defense systems. Defensive AI applications, such as anomaly detection and cyber intrusion systems, require constant innovation and adaptation, demanding significant investment from both the private and public sectors. Governments will need to collaborate with industry leaders to ensure that defensive AI technologies keep pace with the evolving threat landscape. Policymakers should incentivize research into AI that enhances societal resilience to disruptions, focusing on safeguards, detection mechanisms, and regulatory standards that can mitigate potential harms while promoting innovation in AI safety.

This societal scale focus---accounting for the broader sociotechnical ecosystem---can help policymakers and leaders understand and examine how these systems interact with societal norms, legal taxonomies, and global power dynamics. With this macroscopic picture, leaders and policymakers can develop interventions that take into account first and second-order effects to ensure societal resilience.

\section{Future Work}

While this taxonomy offers an initial lens to understand how AI technologies influence societal risk and safety, future research should aim to empirically operationalize it, for example by developing granular quantitative elements for each element. By establishing measurable indicators for factors such as capability breadth, adaptability, and accessibility, the taxonomy's utility for policymakers and practitioners can be significantly enhanced. These metrics would support more precise risk assessments and inform critical decisions related to the development, deployment, and regulation of AI technologies.

A promising method to operationalize the taxonomy is through the application of graph theory and hypergraphs to fully map the complex interdependencies between its elements. By representing key factors like raw capability potential, adaptability, and sociotechnical context as nodes and edges in a network, researchers can visualize relationships that illustrate how these elements influence offense-defense balances. This graphical representation allows for a clearer understanding of the intricate connections between factors. Using computational models based on these graphs, researchers can simulate different scenarios, predicting how changes in one element affect the overall dynamics. This approach enables the identification of critical nodes or connections---leverage points---where interventions could most effectively shift the balance toward defensive advantages.

In addition, synthetic agent-based modeling presents a valuable method for examining the dynamics of offense and defense within AI systems in controlled environments. By creating virtual settings where simulated agents with offensive or defensive capabilities interact, researchers can observe emergent behaviors and identify conditions that lead to either offensive dominance or defensive resilience. This approach allows for experimentation with various scenarios and parameters, offering insights into how specific factors within the Framework affect real-world outcomes. Such modeling can help anticipate potential threats and evaluate the effectiveness of different defensive strategies before they manifest in practice.

Finally, empirical case studies examining real-world instances of AI use and misuse would help validate the taxonomy's applicability and highlight areas for refinement. These studies could provide practical examples of how offense-defense dynamics manifest in various contexts, offering valuable lessons for policymakers and developers alike.

\section{Conclusions}

The advancement of artificial intelligence presents great opportunities for societal benefit alongside significant risks of harm. This paper introduced the a taxonomy of considerations shaping offense-defense dynamics as a means to dissect and understand the factors influencing whether AI technologies contribute more to societal safety or risk.

By defining key concepts such as Raw Capability Potential, Accessibility and Control, Adaptability, Proliferation, Diffusion, and Release Methods, Safeguards and Mitigations, and Sociotechnical Context, we provide an initial structured approach to analyze the potential impacts of AI systems. This sociotechnically informed framing underscores that while AI technologies may be inherently neutral, their impact is heavily influenced by how they are developed, deployed, and governed.

The implications for AI governance and policy are profound. Policymakers must navigate the delicate balance between fostering innovation and ensuring security, promoting defensive applications while curbing offensive uses. Future work is essential to refine this taxonomy and enhance its applicability, and empirical validation and enhanced modeling will be critical in advancing our understanding of offense-defense dynamics. An emerging Offense-Defense Dynamics Framework may serve as a guide for developing nuanced policies that consider the complex nature of these potent technologies.

\bibliographystyle{plainnat}
\bibliography{references}

\end{document}